\documentclass[runningheads]{llncs}
\usepackage{graphicx}
\usepackage{comment}
\usepackage{wrapfig}
\usepackage{amsmath,amssymb} 
\usepackage{color}

\usepackage{paralist}
\usepackage{times}
\usepackage{mfirstuc}
\usepackage{soul}
\usepackage{epsfig}
\usepackage{cite}
\usepackage{multirow}
\usepackage{diagbox}
\usepackage{rotating}
\usepackage{overpic}
\usepackage{booktabs}
\usepackage{array}
\usepackage{colortbl}
\usepackage{tasks}

\newcommand{\nameofblock}{{{sandglass block}}}
\newcommand{\nameofnet}{{{MobileNeXt}}}

\def\ie{\emph{i.e.,\ }}
\def\eg{\emph{e.g.,\ }}

\usepackage{array}
\newcolumntype{x}[1]{>{\centering\arraybackslash}p{#1}}

\usepackage{pifont}

\newcommand{\myPara}[1]{\vspace{5pt}\noindent\textbf{#1}}

\graphicspath{{./figs/}}

\newcommand{\addFig}[1]{}
\newcommand{\addFigs}[1]{}

\usepackage[pagebackref=true,letterpaper=true,colorlinks=true,linkcolor=black,citecolor=black,bookmarks=false]{hyperref}

\begin{document}
\pagestyle{headings}
\mainmatter
\def\ECCVSubNumber{1101}  



\title{Rethinking Bottleneck Structure   for   Efficient \\ Mobile Network Design} 

\makeatletter
\renewcommand*{\@fnsymbol}[1]{\ensuremath{\ifcase#1\or *\or \dagger\or \ddagger\or
    \mathsection\or \mathparagraph\or \|\or **\or \dagger\dagger
    \or \ddagger\ddagger \else\@ctrerr\fi}}
\makeatother

\titlerunning{Rethinking Bottleneck Structure for   Efficient Mobile Network Design}
%
\author{Daquan Zhou\inst{1,2}\thanks{Authors contributed equally.}\thanks{Work done during an internship at Yitu Tech.} \and
Qibin Hou\inst{1}$^*$ \and
Yunpeng Chen\inst{2} \and Jiashi Feng\inst{1} \and Shuicheng Yan\inst{2}}
\authorrunning{D. Zhou and Q. Hou et al.}
%
\institute{National University of Singapore, Singapore \and
Yitu Technology \\
\email{\{zhoudaquan21,andrewhoux\}@gmail.com, elefjia@nus.edu.sg, \{yunpeng.chen, shuicheng.yan\}@yitu-inc.com}}


\maketitle

\begin{abstract}
The inverted residual block is dominating architecture design for mobile networks recently. 
It changes the classic residual bottleneck by introducing two design rules:  learning inverted residuals and 
using linear bottlenecks.
In this paper, we rethink the necessity of such design changes and find it may bring risks of  information loss and gradient confusion.
We thus propose to flip the structure and present a novel
bottleneck design, called the \nameofblock{}, that performs identity mapping and spatial transformation at higher dimensions and thus alleviates information loss and gradient confusion effectively.  
Extensive experiments demonstrate that, different from the common belief, such bottleneck
structure is more beneficial than the inverted ones for mobile networks. 
In ImageNet classification, by simply replacing the inverted residual block with our \nameofblock{} without increasing parameters and computation, 
the classification accuracy can be improved by more than 1.7\% over MobileNetV2. 
On Pascal VOC 2007 test set, we observe that there is also 0.9\% mAP improvement in object detection.
We further verify the effectiveness of the \nameofblock{} by adding it into the search space of neural architecture search method DARTS.
With 25\% parameter reduction, the classification accuracy is improved by 0.13\% over previous DARTS models. 
Code can be found at: \url{https://github.com/zhoudaquan/rethinking\_bottleneck\_design}.

\end{abstract}

\section{Introduction} \label{sec:introduction}
\begin{figure}[t]
    \centering
    \begin{minipage}[c]{0.6\textwidth}
    \tiny
    \begin{overpic}[width=\textwidth]{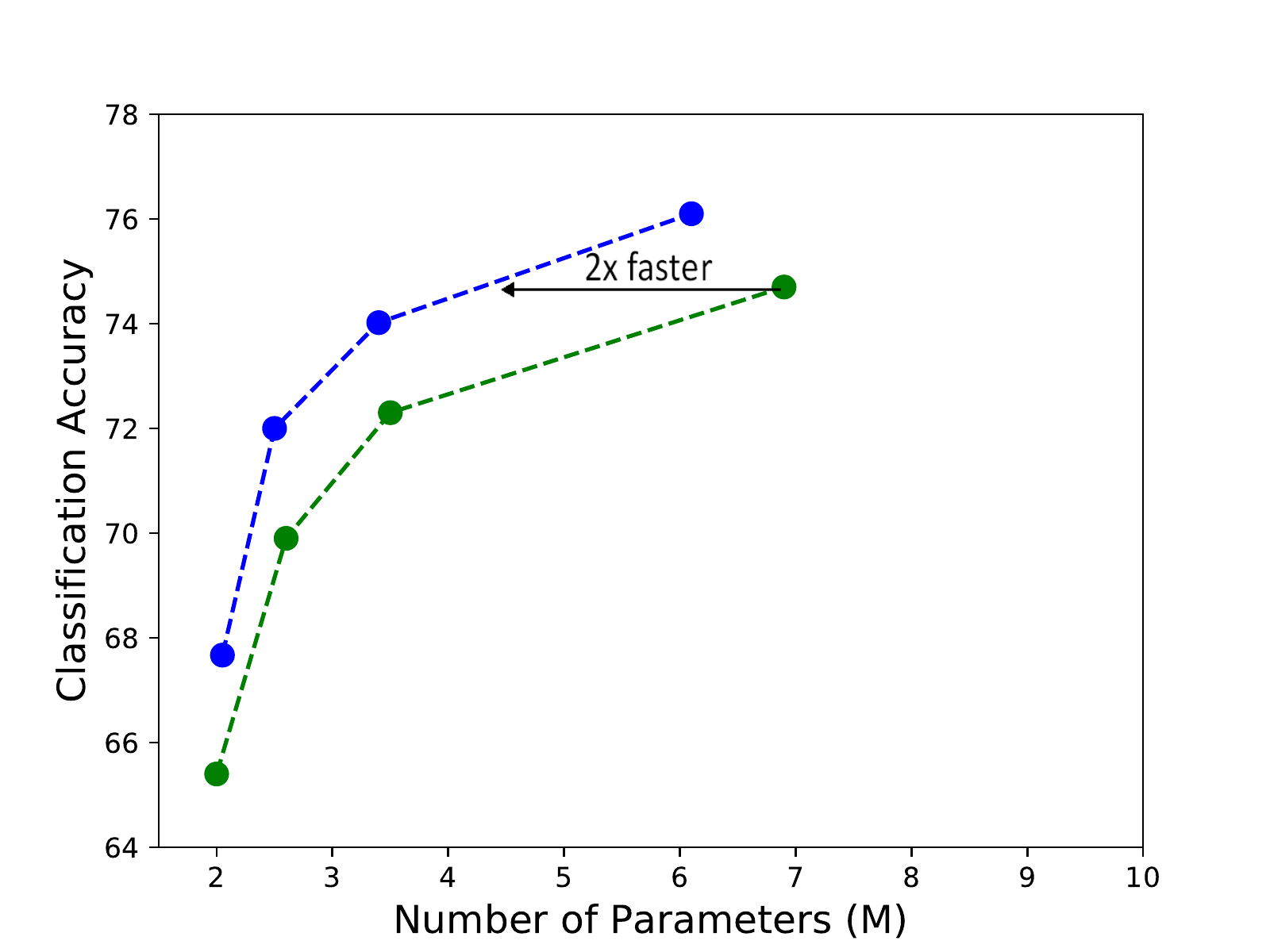}
    \put(64,51){MobileNetV2-1.4}
    \put(32,40){MobileNetV2-1.0}
    \put(24,32){MobileNetV2-0.75}
    \put(19,13.2){MobileNetV2-0.5}
    \put(56,57){MobileNeXt-1.4}
    \put(22,51.5){MobileNeXt-1.0}
    \put(13,38){MobileNeXt-0.75}
    \put(19,23){MobileNeXt-0.5}
    \put(41,23){
    \tiny
    \setlength\tabcolsep{0.5mm}
    \renewcommand{\arraystretch}{1}
    \begin{tabular}{l|ccc} 
      Method  & \#Param. & M-adds & Acc.  \\ \hline
      MobileNetV2-1.4  & 6.9M & 690M & 74.9\%\\
      MobileNeXt-1.4  & 6.1M & 590M & 76.1\%\\ \hline
      MobileNetV2-1.0  & 3.5M & 300M & 72.3\%\\
      MobileNeXt-1.0 &3.4M & 300M &74.0\%\\ \hline
      MobileNetV2-0.75 &2.6M  & 150M & 69.9\%\\
      MobileNeXt-0.75 &2.5M & 210M & 72.0\% \\ \hline
      MobileNetV2-0.5 &1.7M  & 97M & 65.4\%\\
      MobileNeXt-0.5 &1.8M & 110M & 67.7\%\\
    \end{tabular}}
    \end{overpic}
    \end{minipage}\hfill
    \begin{minipage}[c]{0.4\textwidth}
    \caption{
    Top-1 classification accuracy comparisons between the proposed \nameofnet{} and MobileNetV2 \cite{sandler2018mobilenetv2}.
    We use different width multipliers to trade-off between model complexity and accuracy. Here, four widely-used multipliers are chosen, including 0.5, 0.75, 1.0, and 1.4. As can be seen, under each width multiplier, our \nameofnet{} surpasses the MobileNetV2 
    baseline by a large margin, especially for the models with less learnable parameters. 
    }
    \end{minipage}
	\label{fig:imagenet}
\end{figure}
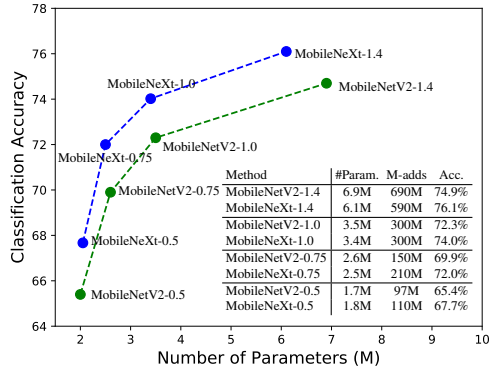
\begin{figure}[t]
    \centering
    \includegraphics[width=0.9\linewidth]{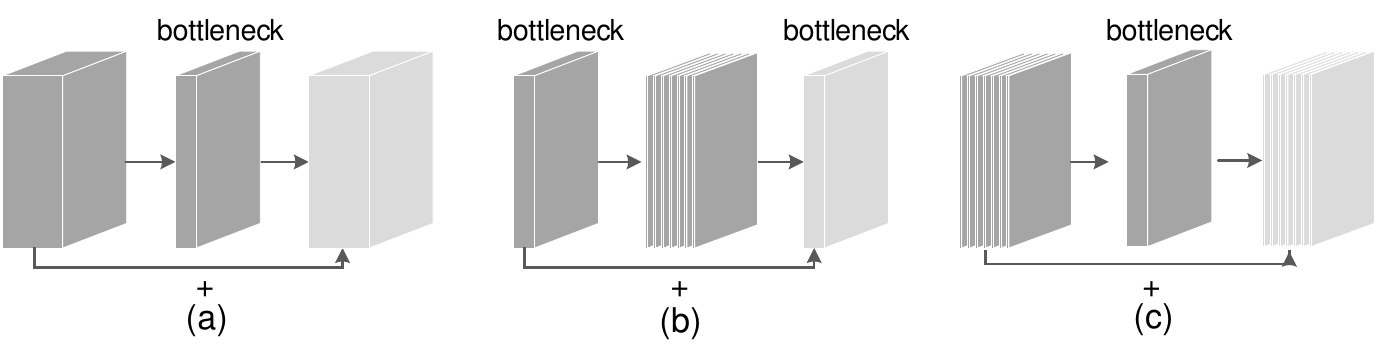}
    \caption{
    Conceptual diagram of different residual bottleneck blocks. 
    (a) Classic residual block with bottleneck structure \cite{He2016}. 
    (b) Inverted residual block  \cite{sandler2018mobilenetv2}.
    (c) Our proposed \nameofblock{}. 
    We use thickness of each block to represent the corresponding relative number of channels.
    As can be seen, compared to the inverted residual block, the proposed residual block reverses the
    thought of building shortcuts between bottlenecks
    and adds depthwise convolutions (detached blocks) at both ends of the residual path, both of which are found crucial for performance improvement. 
    }
    \label{fig:comps}
\end{figure}

A common belief behind the design principles of most popular light-weight models (either manually designed or automatically searched) \cite{sandler2018mobilenetv2,ma2018shufflenet,tan2019mnasnet,tan2019efficientnet}
is to adopt the inverted residual block~\cite{sandler2018mobilenetv2}.
Compared to the classic residual bottleneck block~\cite{he2016deep,he2016identity}, this block shifts the identity mapping from high-dimensional representations to low-dimensional ones (\ie the bottlenecks).
However, connecting identity mapping between thin bottlenecks
would inevitably lead to information loss since the residual
representations are compressed as shown in Figure~\ref{fig:comps}(b).
Moreover, it would also weaken the propagation capability of gradients across layers, due to gradient confusion arising from the narrowed feature dimensions, and hence affect the training convergence and model performance~\cite{sankararaman2019impact}. 
Therefore, despite the wide use of the inverted residual block,
how to design residual blocks for mobile devices is very worthy of
studying.

In this paper, in view of the above concerns, we rethink the rationality of shifting from the classic bottleneck structure (Figure~\ref{fig:comps}(a)) to the popular inverted residual block (Figure~\ref{fig:comps}(b)) in developing mobile networks. 
In particular, we consider the following three fundamental  questions.
(i) What are the effects if we position the identity mapping (\ie shortcuts) at the high-dimensional representations as done in the classic bottleneck structure?
(ii) While the linear activation can reduce information loss, should it only be applied to the bottlenecks? 
(iii) The previous questions remind us of the classic bottleneck structure which suffers high computational complexity. 
This cost can be reduced by replacing the dense spatial convolutions with depthwise ones, but, regarding the bottlenecks, should the depthwise convolution be still added in the low-dimensional bottleneck as conventional?

Motivated by the above questions, we present and evaluate a new bottleneck design, termed the  \nameofblock{}.
Unlike the inverted residual block that builds shortcuts between linear bottlenecks, our \nameofblock{}  puts shortcut connections between linear high-dimensional representations, as shown in Figure~\ref{fig:comps}(c).
Such structure preserves more information delivered
between blocks compared to the inverted residual block and propagates more gradients backward to better optimize network training because of the high-dimensional residuals~\cite{sankararaman2019impact}.
Furthermore, to learn more expressive spatial representation, 
instead of putting the spatial convolutions in the bottleneck with compressed channels,
we propose to apply them in the expanded high dimensional feature space,
which we find is an effective way of improving the model performance.
In addition, we maintain the channel reduction and expansion process
with pointwise convolutions to reduce computational cost.
This makes our block quite different from the inverted residual block but more similar to the classic residual bottleneck.

We stack the \nameofblock{}s in a modularized way to build the proposed \nameofnet{}. 
Our network achieves more than 1.7\%  top-1 classification accuracy improvement over MobileNetV2 on ImageNet with slightly less computation and a comparable number of parameters as shown in Figure~\ref{fig:imagenet}. 
When applying the \nameofblock{} on the EfficientNet topology to replace the inverted residual block, the resulting model surpasses the previous state-of-the-art by 0.5\% with a comparable amount of computation but 20\% parameter reduction.
Particularly, in object detection, when taking SSDLite~\cite{liu2016ssd,sandler2018mobilenetv2}
as the object detector, using our \nameofnet{} as backbone gains 0.9\% in mAP on the Pascal VOC 2007 test set over MobileNetV2.
More interestingly, we also experimentally find the proposed \nameofblock{} can be used to enrich the search space of neural architecture search algorithms~\cite{liu2018darts}.
By adding the \nameofblock{} into the search space as a `super' operator, 
without changing the search algorithm, the resultant model can improve classification accuracy by 0.13\% but with 25\% less parameters compared to models searched from the vanilla space.

In summary, we make the following contributions in this paper:

\begin{itemize}

     \item {Our results advocate a rethinking of the bottleneck structure for mobile network design. It seems that     the inverted residuals  are not so advantageous over the bottleneck structure as commonly believed.}
     
     \item 
     Our study reveals that building shortcut connections along higher-dimensional feature space could promote model performance. 
     Moreover, depthwise convolutions should be conducted in the high dimensional space for 
     learning more expressive features and learning linear residuals is also crucial for bottleneck structure.
     
    \item Based on our study, we propose a novel  \nameofblock{}, which substantially extends the classic bottleneck structure. 
    We experimentally demonstrate that this structure is more suitable for mobile applications in terms of both accuracy and efficiency and 
    can be used as `super' operators in architecture search algorithms for better  architecture generation.

     
\end{itemize}


\section{Related Work}

Modern deep neural networks are mostly built by stacking building blocks, which are designed based on either the classic residual block with bottleneck structure~\cite{he2016deep} or the inverted residual block~\cite{sandler2018mobilenetv2}.
%
In this section, we categorize all related networks based on above two types of building blocks and briefly describe them below.

\myPara{Classic residual bottleneck blocks}
The bottleneck structure was first introduced in
ResNet~\cite{he2016deep}.
A typical bottleneck structure consists of three convolutional layers: an $1\times 1$ convolution for channel reduction, 
a $3\times3$ convolution for spatial feature extraction, and
another $1\times 1$ convolution for channel expansion.
A residual network is often constructed by stacking a sequence of such residual blocks. 
%
%
The bottleneck structure was further developed in later works by widening the channels in each convolutional layer~\cite{zagoruyko2016wide}, applying group convolutions to the middle
bottleneck convolution for aggregating richer feature representations~\cite{xie2017aggregated},  or introducing attention based modules
to explicitly model inter-dependencies between channels~\cite{hu2018squeeze,li2019selective}.
%
There are also other works \cite{chen2017dual,touvron2019fixing} 
combining residual blocks with dense connections to boost the performance. 
%
However, in spite of the success in heavy-weight network design,
it is rarely used in light-weight networks due to the model complexity.
Our work demonstrates that by reasonably adjusting the residual block, this kind of classic bottleneck structure is also suitable for light-weight networks and can yield state-of-the-art results.


\myPara{Inverted residual blocks} The inverted residual block, which was first introduced in MobileNetV2~\cite{sandler2018mobilenetv2}, reverses the idea of the classic bottleneck structure and connects shortcuts between linear bottlenecks.
It largely improves performance and optimizes the model complexity compared to the classic 
MobileNet~\cite{howard2017mobilenets} which is composed of a sequence of $3\times3$ depthwise separable convolutions.
Because of high efficiency, the inverted residual block has been widely adopted in the later mobile network architectures. 
ShuffleNetV2~\cite{ma2018shufflenet} inserts a channel split module before the inverted residual block and adds another channel shuffle module after it.
In HBONet~\cite{li2019hbonet}, down-sampling operations are introduced into inverted residual blocks for modeling richer spatial information.
%
%
MobileNetV3~\cite{howard2019searching} proposes to search for optimal activation functions and the expansion rate of inverted residual blocks at each stage.
More recently, MixNet~\cite{tan2019mixconv} proposes to search for optimal kernel sizes of the depthwise separable convolutions in the inverted residual block. 
EfficientNet~\cite{tan2019efficientnet} is also based on the inverted residual block but differently it uses a scaling method to control the network weight in terms of input resolution,
network depth, and network width.
%
%
Different from all the above approaches, our work advances the standard bottleneck structure
and demonstrates the superiority of our building block over the inverted residual block in mobile settings.

\myPara{Model compression and neural architecture search}
Model compression algorithms are effective for removing redundant parameters for neural networks, such as network pruning \cite{liu2017learning,han2015deep,radu2019performance,caron2020pruning}, quantization \cite{hubara2017quantized,choukroun2019low}, factorization \cite{zhou2019tensor,jaderberg2014speeding}, and knowledge distillation \cite{hinton2015distilling}. 
Despite efficient networks, the performance of the 
compressed networks is still closely related to the
original networks' architectures. 
Thus, designing more efficient network architectures is essential for yielding efficient models.
%
%
Neural architecture search achieves so by automatically searching efficient network architectures~\cite{tan2019mnasnet,cai2018proxylessnas,guo2019single}. 
However, the search space requires human expertise and the performance of the searched networks is largely dependent upon the designed search space as pointed out in~\cite{ying2019bench, dong2020bench}. In this paper, we show that our proposed building block is complementary to existing search space design principles and can further improve the performance of searched networks if added to existing search spaces.


\section{Method} \label{sec:method}

In this section, we first review some preliminaries about the bottleneck structure widely used in  previous residual networks 
and then describe our proposed  \nameofblock{} and 
network architecture. 
%

%
\subsection{Preliminaries}

\myPara{Residual block with bottleneck structure}
%
%
The classic residual block with bottleneck structure~\cite{he2016deep},  
as shown in Figure~\ref{fig:comps}(a), consists of  two $1\times1$ convolution layers
for channel reduction and expansion respectively and one  $3\times3$ convolution layer between them   for spatial information encoding.
%
In spite of its success in heavy-weight network design~\cite{he2016deep},
this conventional bottleneck structure is not suitable for building 
light-weight neural networks because of its large amount of parameters 
and computation cost in the standard $3\times3$ convolutional layer. 

\myPara{Depthwise separable convolutions}
To reduce computational cost and make the network more efficient, depthwise separable convolutions~\cite{chollet2017xception,howard2017mobilenets} are developed 
to replace the standard one.
As demonstrated in~\cite{chollet2017xception}, a convolution with a $k \times k \times M \times N$
weight tensor,  where $k \times k$ is the kernel size and $M$ and $N$
are the number of input and output channels respectively, 
can be factorized into two convolutions. 
The first is an $M$-channel $k \times k$ depthwise (\emph{a.k.a} channel-wise)
convolution to learn the spatial correlations among
locations within each channel separately. 
The second is a pointwise convolution
that learns to linearly combine channels to 
produce new features.
%
%
%
As the combination of a pointwise convolution and a 
$k \times k$ depthwise convolution has significantly 
less parameters and computations, 
using depthwise separable convolutions in basic building blocks can remarkably reduce the parameters and computational cost. Our proposed architecture also adopts such separable convolutions.
%

\myPara{Inverted residual block} 
The inverted residual block is specifically tailored for mobile devices, especially
those with limited computational resource budget.
%
%
More specifically, unlike the classic bottleneck structure
as shown in Figure~\ref{fig:comps}(a), to save computations, 
it takes as input a low-dimensional compressed
tensor and expands it to a higher dimensional one  by a
pointwise convolution. Then it applies depthwise 
convolution for spatial context encoding, followed by another
pointwise convolution to generate a low-dimensional feature
tensor as input to the next block.
The inverted residual block presents two distinct architecture designs for gaining efficiency without suffering too much performance drop:  the shortcut connection is put between the low-dimensional bottlenecks
if necessary (as shown in Figure~\ref{fig:comps}(b)); and linear bottleneck is adopted.

Despite good performance~\cite{sandler2018mobilenetv2},
in inverted residual blocks, feature maps encoded 
by the intermediate expansion layer should be
first projected to low-dimensional ones, which may not preserve enough useful information
due to channel compression. 
%
%
Moreover, recent studies have unveiled that wider architecture is more favorable for alleviating gradient
confusion~\cite{sankararaman2019impact} and hence can improve
network performance. 
Putting shortcut connections between bottlenecks may prevent 
the gradients from top layers from being successfully propagated to bottom layers during model training because of the  low-dimensionality of representations between adjacent   inverted residual blocks.  

\subsection{Sandglass Block}

\label{subsess:arch_design}

In view of the aforementioned limitations of the inverted
residual block,  we rethink its design rules and present a
\nameofblock{} that can tackle the above issues
by flipping the thought of inverted residuals.

Our design principle is mainly based on the following insights:
(i) To preserve more information from the bottom layers when  transiting to the top layers and to facilitate  the  
gradients propagation across layers, the shortcuts should be 
positioned to connect high-dimensional representations.
(ii) Depthwise convolutions with small kernel size (\eg $3\times3$)
are light-weight, so we can appropriately apply a couple of depthwise convolutions onto the higher-dimensional features such that richer spatial information can be encoded to generate more expressive representations. 
%
%
%
We elaborate on these design considerations in the following.

\myPara{Rethinking the positions of expansion and reduction layers} Originally, the inverted residual block performs expansion at first and then reduction.
Based on the aforementioned design principle, 
to make sure the shortcuts connect high-dimensional representations,
we propose to reverse the order of the two pointwise convolutions first.
Let $\mathbf{F} \in \mathbb{R}^{D_f \times D_f \times M}$ 
be the input tensor and $\mathbf{G} \in \mathbb{R}^{D_f \times D_f \times M}$ the output tensor of a building block\footnote{For simplicity, we assume that the input and output of the building block share the same number of channels and resolution.}.
%
%
We do not consider the depthwise convolution and activation layers at this moment. 
The formulation of our building block can be written as follows:
\begin{equation}
\label{eqn:basic_block}
    \mathbf{G} = \phi_{e}(\phi_{r}(\mathbf{F})) + \mathbf{F},
\end{equation}
where $\phi_e$ and $\phi_r$ denote the two pointwise convolutions
for channel expansion and reduction, respectively.
%
In this way, we can keep the bottleneck in the middle of 
the residual path for saving parameters and computation cost.
More importantly, this allows us to use the shortcut connection to connect representations with a large number of channels instead of the bottleneck ones.

\begin{figure}[t]
    \centering
    \includegraphics[width=0.8\textwidth]{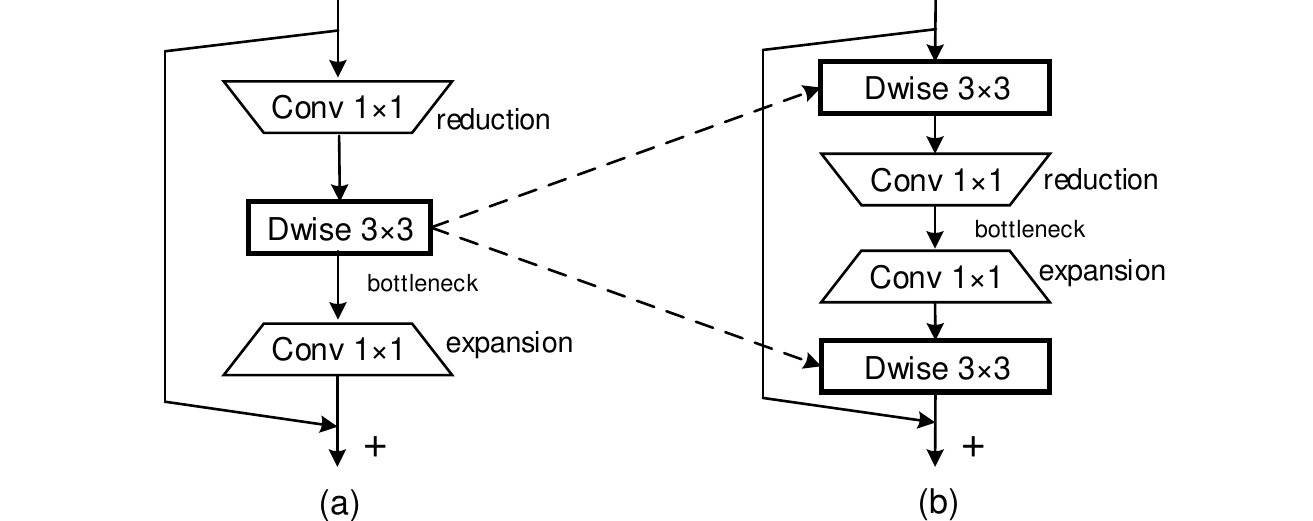}
    \caption{Different types of residual blocks. 
    (a) Classic bottleneck structure with depthwise spatial convolutions.
    (b) Our proposed \nameofblock{} with bottleneck structure.
    To encode more expressive spatial information, instead of adding depthwise convolutions in the bottleneck, we propose to move them to the ends of the residual path, which have high-dimensional representations.
    }
    \label{fig:arch}
\end{figure}

\myPara{High-dimensional shortcuts} Instead of putting the
shortcut between bottlenecks, we put the shortcuts between higher-dimensional representations as shown in Figure~\ref{fig:arch}(b). 
The `wider' shortcut delivers more information from the input
$\mathbf{F}$ to the output $\mathbf{G}$ compared to the
inverted residual block and allows more gradients to propagate across multiple layers.

%

\myPara{Learning expressive spatial features}
Pointwise convolutions can be used to encode the inter-channel information but fail to capture spatial information.
In our building block, we follow previous mobile networks
and adopt depthwise spatial convolutions to encode spatial
information.
The inverted residual block adds depthwise convolutions
between pointwise convolutions to learn expressive spatial
context information.
However, in our case, the position between two pointwise
convolutions is the bottleneck.
Directly adding depthwise convolutions in the bottleneck
as shown in Figure~\ref{fig:arch}(a) makes them have fewer filters and thus, less spatial information can be encoded.
We experimentally find that this structure largely degrades
the performance compared to MobileNetV2 by more than 1\%.

Regarding the positions of the pointwise convolutions,
instead of directly putting the depthwise convolution between
the two pointwise convolutions, we propose to add depthwise
convolutions at the ends of the residual path as shown in 
Figure~\ref{fig:arch}(b).
Mathematically, our building block can be formulated as follows:
\begin{align}
\label{eqn:i2r_block}
    \mathbf{\hat{G}} &= \phi_{1,p}\phi_{1,d}(\mathbf{F})\\
    \mathbf{G} &= \phi_{2,d}\phi_{2,p}(\mathbf{\mathbf{\hat{G}}}) + \mathbf{F}
\end{align}
%
where $\phi_{i,p}$ and $\phi_{i,d}$ are the $i$-th pointwise
convolution and depthwise convolution, respectively.
In this way, since both depthwise convolutions
are conducted in high-dimensional spaces, richer feature 
representations can be extracted compared to the inverted
residual block.
We will give more explanations on the advantages of
such design.

\begin{table}[t]
    \centering
    \small
    \setlength\tabcolsep{5.8mm}
    \renewcommand{\arraystretch}{1.1}
         \caption{Basic operator description of the proposed 
         \nameofblock{}. Here, `t' and `s' denote the channel
         reduction ratio and the stride, respectively.}
    \begin{tabular}{cccccccc} \toprule[1pt] 
        Input dimension & Operator type & Output dimension    \\ \midrule[1pt]
        $D_f \times D_f \times M$ & 3$\times$3 Dwise conv, ReLU6 & $D_f \times D_f \times M$ \\ 
        $D_f \times D_f \times M$ & 1$\times$1 conv, linear & $D_f \times D_f \times \frac{M}{t}$  \\ 
        $D_f \times D_f \times \frac{M}{t}$  & 1$\times$1 conv, ReLU6 & $D_f \times D_f \times N$  \\
        $D_f \times D_f \times N$ & 3$\times$3 Dwise conv, linear, stride = $s$ &  $\frac{D_f}{s} \times \frac{D_f}{s} \times N$ \\ 
        \bottomrule[1pt]
        \end{tabular}
        \label{tab:block}
\end{table}

\myPara{Activation layers}
%
It has been demonstrated in~\cite{sandler2018mobilenetv2} that using linear bottlenecks can help prevent the feature values from being zeroed and hence reduce information loss.
Following this suggestion, we do not add any activation layer
after the reduction layer (the first pointwise convolutional layer).
It should also be noted that though the output of our
building block is high-dimensional, we empirically find
adding an activation layer after the last convolution can 
negatively influence the classification performance.
Therefore, activation layers are only added after the first
depthwise convolutional layer and the last pointwise convolutional layer.
We will give more explanations in our experiments on this.

\myPara{Block structure}
Taking the above considerations, we design a novel residual bottleneck block.
The structure details are given in Table~\ref{tab:block}, and the diagram can also be found in Figure~\ref{fig:arch}(b).
Note that when the input and output have different channel numbers, we do not add the shortcut connection.
For depthwise convolutions, we always use 
kernel size $3\times3$ as done in other works~\cite{he2016deep,sandler2018mobilenetv2}.
We also utilize batch normalization and ReLU6
activation if necessary during training.

\myPara{Relation to the inverted and classic residual blocks}
Albeit both architectures exploit the bottlenecks,
the design intuition and the internal structure are quite
different.
Our goal is to demonstrate that the idea of building shortcut connections between high-dimensional representations
as in the classic bottleneck structure \cite{he2016deep}
is suitable for light-weight networks as well.
To the best of our knowledge, this is the first work
that attempts to investigate the advantages of the classic
bottleneck structure over the inverted residual block for efficient network design.
%
%
On the other hand, we also attempt to demonstrate that
adding depthwise convolutions to the ends of the residual path in our structure can encourage the network to learn more expressive spatial information and hence yield better performance.
In our experiment section, we will show more numerical results and provide detailed analysis.

\subsection{\nameofnet{} Architecture}
\label{sec:full_arch}
Based on our \nameofblock{}, we develop a modularized architecture, \nameofnet{}.
At the beginning of our network, there is a convolutional layer with 32 output channels.
After that, our \nameofblock{}s are stacked together.
Detailed information about the network architecture can be found in Table~\ref{tab:i2rnet_arch}.
%
%
%
Following \cite{sandler2018mobilenetv2}, the expansion ratio used in our network is set to 6 by default.
The output of the last building block is followed by a global average pooling layer to transform 2D feature maps to 1D feature vectors.
A fully-connected layer is finally added to predict the final score for each category.

\myPara{Identity tensor multiplier}
%
%
The shortcut connections in residual blocks have been shown
essential for improving the capability of propagating gradients across layers~\cite{he2016deep,sandler2018mobilenetv2}.
According to our experiments, we find that there is no need
to keep the whole identity tensor to combine with the residual path.
To make our network more friendly to mobile devices, 
we introduce a new hyper-parameter---identity tensor 
multiplier, denoted by $\alpha \in [0,1]$, which controls 
what portion of the channels in the identity tensor is preserved.
For convenience, let $\phi$ be the transformation function of the residual path in our block.
Originally, the formulation of our block can be written as
$G= \phi(F) + F$.
After applying the multiplier, our building block can be rewritten as 
\begin{equation}
    G_{1:\alpha M} = \phi(F)_{1:\alpha M} + F_{1:\alpha M}, \quad G_{\alpha M:M} = \phi(F)_{\alpha M:M},
\end{equation}
where the subscripts index the channel dimension.

The advantages of using $\alpha$ are mainly two-fold.
First, after reducing the multiplier, the number of
element-wise additions in each building block can be reduced.
As pointed out in~\cite{ma2018shufflenet}, the element-wise
addition is time consuming. 
Users can choose a lower identity tensor multiplier to yield
better latency with nearly no performance drop.
Second, the number of memory access times can be reduced.
One of the main factors that affect the model latency is 
the memory access cost (MAC).
As the shortcut identity tensor is from the output of the last building block, its recurrent nature hints an opportunity to cache it on the chip in order to avoid the excessive off-chip memory access. 
Therefore, reducing the channel dimension of the identity
tensor can effectively encourage the processors to store it 
in the cache or other faster memory near the processors 
and hence improve the latency.
We will give more details on how this multiplier affects
the performance and model latency in the experiment section.

\begin{table}[t]
    \centering
    \small
    \setlength\tabcolsep{3mm}
    \renewcommand{\arraystretch}{1}
     \caption{Architecture details of the proposed MobileNeXt. Each row denotes a sequence of 
        building blocks, which is repeated `b' times. The reduction ratio
        used in each building block is denoted by `t'.
        The stride of the first building block
        in each stage is set to 2 and all the others are with stride 1. Each convolutional
        layer is followed by a batch normalization layer and the kernel size for all
        spatial convolutions is set to $3\times3$. We do not add identity mappings
        for those blocks have different input and output channels. We suppose there are totally $k$ categories.}
    \begin{tabular}{cccccccc} \toprule[1pt] 
        No. & t & Output dimension & s & b & Input dimension & Operator   \\ \midrule[1pt]
        1 & - & $112\times112\times32$ & 2 &  1 & $224\times224\times3$ & conv2d 3x3\\ 
        2 & 2 & $56\times56\times96$ & 2 &  1 & $112\times112\times32$ & \nameofblock{}\\ 
        3 & 6 & $56\times56\times144$ & 1 &  1 & $56\times56\times96$ & \nameofblock{} \\ 
        4 & 6 & $28\times28\times192$ & 2 &  3 & $56\times56\times144$ & \nameofblock{}  \\
        5 & 6 & $14\times14\times288$ & 2 &  3 & $28\times28\times192$ & \nameofblock{}  \\
        6 & 6 & $14\times14\times384$ & 1 &  4 & $14\times14\times288$ & \nameofblock{}  \\
        7 & 6 & $7\times7\times576$ & 2 &  4 & $14\times14\times384$ & \nameofblock{}  \\ 
        8 & 6 & $7\times7\times960$ & 1 &  2 & $7\times7\times576$ & \nameofblock{} \\ 
        9 & 6 & $7 \times 7 \times 1280$ & 1 & 1 & $7\times7\times960$ & \nameofblock{} \\
        10 & - & $1\times1\times1280$ & - & 1 & $7 \times 7 \times 1280$ & avgpool 7x7  \\
        11 & - & $k$ & - &  1 & $1\times1\times1280$ & conv2d 1x1\\ 
        \bottomrule[1pt]
        \end{tabular}
        \label{tab:i2rnet_arch}
\end{table}

\section{Experiments}

\subsection{Experiment Setup}

We adopt the PyTorch toolbox \cite{paszke2019pytorch} to
implement all our experiments.
We use the standard SGD optimizer to train our models
with both decay and momentum of 0.9 and the weight decay is $4\times 10^{-5}$.
We use the cosine learning schedule with an initial learning rate of 0.05.
The batch size is set to 256 and four GPUs are used for training.
Without special declaration, we train all the models for 200 epochs and report results on the ImageNet \cite{krizhevsky2012imagenet} 
for classification and Pascal VOC dataset \cite{pascal-voc-2012} 
for object detection. We use distributed training with three epochs of warmup.

\subsection{Comparisons with MobileNetV2}
\label{subsec:comp_mbv2}
In this subsection, we extensively study the advantages of our \nameofnet{}{}  over MobileNetV2 under various settings.
Besides comparing performance of their full models (\ie, with weight multiplier of 1) for classification, we also compare their performance with other weight multipliers and  quantization. This can help unveil the performance advantage of our model w.r.t.\ the full spectrum of model architecture configurations.

\myPara{Comparison under different width multipliers}
We use the width multiplier as a scaling factor to trade off the model complexity and accuracy of the model as used in \cite{howard2017mobilenets,sandler2018mobilenetv2,howard2019searching}.
%
Here, we adopt five different multipliers, 
including 1.4, 1.0, 0.75, 0.5, and 0.35, to show the superiority of our network over MobileNetV2.
%
As can be seen in Table~\ref{tab:i2rnet_imagenet_224} \footnote{We also conduct latency measurements with TF-Lite on Pixel 4XL and the measured latency for \nameofnet{} and MobileNetV2 are 66ms and 68 ms respectively.},
our networks with different multipliers all
outperform MobileNetV2 with comparable numbers of learnable parameters and computational cost. 
The performance gain of our model over MobileNetV2 is especially high when the multiplier is small. 
This demonstrates that our model is more
efficient since our model performance is much better
at small sizes.


\begin{table}[t]
    \centering
    \small
    \setlength\tabcolsep{3mm}
    \renewcommand{\arraystretch}{1.1}
    \caption{Comparisons with MobileNetV2 using different width multipliers with input resolution $224\times224$. As can be seen, the smaller the multiplier is set to the better performance gain we achieve over MobileNetV2
    with comparable latency (\eg 210ms  for both models with width multiplier 1.0) tested on Google Pixel 4XL under the PyTorch environment setting.} 
    \begin{tabular}{cccccccc} \toprule[1pt] 
        No. & Models & Param. (M) & MAdd(M) & Top-1 Acc. (\%)  \\ \midrule[1pt]
        1 & MobileNetV2-1.40 & 6.9 & 690 &  74.9  \\
        2 & MobileNetV2-1.00 & 3.5 & 300 &  72.3  \\ 
        3 & MobileNetV2-0.75 & 2.6 & 150 &  69.9  \\ 
        4 & MobileNetV2-0.50 & 2.0 & 97 &  65.4  \\
        5 & MobileNetV2-0.35 & 1.7 & 59 &  60.3  \\ \toprule[0.6pt]
        6 & \nameofnet{}{}-1.40 & 6.1 & 590 &  76.1   \\
        7 & \nameofnet{}{}-1.00 & 3.4 & 300 &  74.0   \\
        8 & \nameofnet{}{}-0.75 & 2.5 & 210 &  72.0   \\ 
        9 & \nameofnet{}{}-0.50 & 2.1 & 110 &  67.7   \\
        10 & \nameofnet{}{}-0.35 & 1.8 & 80 &  64.7   \\
        \bottomrule[1pt]
        \end{tabular}
        
        \label{tab:i2rnet_imagenet_224}
\end{table}

\begin{table}[t]
    \centering
    \small
    \setlength\tabcolsep{1.4mm}
    \renewcommand{\arraystretch}{1.1}
     \caption{Performance of our proposed \nameofnet{}{} and MobileNetV2 after post-training quantization. In bites configurations, `W' denotes the number of bits used to represent the weights of the model and `A' denotes the number of bits used to represent the activations.}
    \begin{tabular}{cccccccc} \toprule[1pt] 
        Model & Precision~(W/A)  & Method & Top-1 Acc. (\%)  \\ \midrule[1pt]
        MobileNetV2 & INT8/INT8  & Post Training Quant. &  65.07 \\ 
        \nameofnet{}{} & INT8/INT8  & Post Training Quant. &  $68.62_{+3.55}$   \\\midrule[0.6pt]
        MobileNetV2 & FP32/FP32  & - & 72.25  \\ 
        \nameofnet{}{} & FP32/FP32  & - & $74.02_{+1.77}$ \\
        \bottomrule[1pt]
        \end{tabular}
       
        \label{tab:quantization}
\end{table}

\myPara{Comparison under post-training quantization}
Quantization algorithms are often used in real-world applications as a kind of effective compression tool with subtle performance loss. However, the performance of the quantized model is significantly affected by the original base model.
We experimentally show that the \nameofnet{}{} can achieve better performance than the MobileNetV2 when combined with the quantization algorithm.
Here, we use a widely-used post-training linear quantization method 
introduced in \cite{migacz2017nvidia}.
We apply 8-bit quantization on both weights and activations as 8-bit is the most common scheme used on hardware platforms.
The results are shown in Table~\ref{tab:quantization}.
Without quantization, our network improves MobileNetV2 by more than 1.7\% in terms of top-1 accuracy.
When the parameters and activations are quantized to 8 bits,
our network outperforms MobileNetV2 by 3.55\% under the same
quantization settings.
The reasons for this large improvement are two-fold.
First, compared to MobileNetV2, we move the shortcut in each
building block from low-dimensional representations to high-dimensional ones.
After quantization, more informative feature representations can be preserved.
Second, using more depthwise spatial convolutions can help
preserve more spatial information, which we believe is beneficial to the classification performance.

\begin{table}[t]
    \centering
    \small
    \setlength\tabcolsep{1.4mm}
    \renewcommand{\arraystretch}{1.1}
    \caption{Performance of our proposed network and MobileNetV2 when adding the number of spatial convolutions (Dwise convs) in each building block. 
    Obviously, our \nameofnet{} performs much better than the improved MobileNetV2 with less learnable parameters and computational cost.}
    \begin{tabular}{cccccccc} \toprule[1pt] 
        Method & \#Dwise convs & Param. (M) & M-Adds (M) & Top-1 Acc. (\%)  \\ \midrule[1pt]
        MobileNetV2 & 2 (middle) & 3.6 & 340 &  73.02 \\ 
        \nameofnet{}{} & 2 (top, bottom)  & 3.5 & 300 &  74.02  \\ 
        \bottomrule[1pt]
        \end{tabular}
       
        \label{tab:spatial_conv}
\end{table}

\myPara{Comparison with MobileNetV2 on structure}
As shown in Figure~\ref{fig:arch}(b), our \nameofblock{} contains two $3\times3$ depthwise convolutions for encoding rich spatial context information.
To demonstrate the benefit of our model comes from our
novel architecture rather than leveraging one more depthwise 
convolution or larger receptive field, in this experiment, 
we attempt to compare with an improved version of MobileNetV2 with one more depthwise convolution inserted in the middle of each inverted residual block.
The results are shown in Table~\ref{tab:spatial_conv}.
Obviously, after adding one more depthwise convolution, the performance of MobileNetV2 increases to 73\%, which is still far worse than ours (74\%) with even more learnable parameters and complexity.
This indicates that structurally our network does have an edge over MobileNetV2.



\subsection{Comparison with State-of-the-Art Mobile Networks}

To further verify the superiority of our proposed \nameofblock{} over the inverted residual blocks, we add squeeze and excite modules into our \nameofnet{} as done in \cite{howard2019searching,tan2019efficientnet}.
We do not apply any searching algorithms on the architecture design and data augmentation policy. 
We directly take the EfficientNet-b0 architecture \cite{tan2019efficientnet}
and replace the inverted residual block with \nameofblock{} with the basic augmentation policy. 
As shown in table \ref{tab:cross_comparisons}, 
with a comparable amount of computation and ~20\% parameter reduction, replacing the inverted residual block with 
\nameofblock{} results in 0.4\% top-1  classification accuracy improvement on ImageNet-1k dataset.

\subsection{Ablation Studies}
In Sec.~\ref{subsec:comp_mbv2}, we have shown the importance
of connecting high-dimensional representations with shortcuts.
In this subsection, we study how other model design
choices contribute to the model performance and efficiency,
including the effect of using wider transformation,
the importance of learning linear residuals, 
and the role of identity tensor multiplier.

\myPara{Importance of using wider transformation}
As described in Sec.~\ref{sec:method}, we apply spatial
transformation and shortcut connections to high-dimensional
representations.
To demonstrate the importance of such operations,
we follow the inverted residual block to use the shortcuts
to connect the bottleneck representations.
This operation leads to an accuracy decrease of 1\%,
which indicates applying shortcuts at wider dimension
is more beneficial.

\myPara{Importance of linear residuals}
According to MobileNetV2 \cite{sandler2018mobilenetv2},
its classification performance will be degraded when
replacing the linear bottleneck with the non-linear one because of information loss.
From our experiment, we obtain a more general
conclusion.
We find that though the shortcuts connect high-dimensional
representations in our model, adding non-linear activations (ReLU6) to the last convolutional layer decreases the performance
by nearly 1\% compared to the setting using linear
activations (no ReLU6).
This indicates that learning linear residual (\ie adding no
non-linear activation layer on the top of the residual path) is essential for light-weight networks with shortcuts connecting either expansion layers or reduction layers.

\begin{table}[t]
    \centering
    \small
    \setlength\tabcolsep{3mm}
    \renewcommand{\arraystretch}{1.1}
     \caption{Comparisons with other state-of-the-art models. \nameofnet{}{} denotes the model based on our proposed \nameofblock{} and \nameofnet{}{}$^\dagger$ denotes the models with \nameofblock{} and the SE module \cite{hu2018squeeze} added for a fair comparison with other state-of-the-art models such as EfficientNet. We do not apply any searching algorithms on both the architecture design and data augmentation policy.}
    \begin{tabular}{cccccccc} \toprule[1pt] 
         Models & Param. (M) & MAdd (M) & Top-1 Acc. (\%)    \\ \midrule[1pt]
         MobilenetV1-1.0\cite{howard2017mobilenets} & 4.2 & 575 &   70.6  \\
         ShuffleNetV2-1.5\cite{ma2018shufflenet} & 3.5 & 299 &  72.6\\
         MobilenetV2-1.0\cite{sandler2018mobilenetv2} & 3.5 & 300 &   72.3  \\
         MnasNet-A1\cite{tan2019mnasnet} & 3.9 & 312 &  75.2  \\ 
         MobilenetV3-L-0.75\cite{howard2019searching} & 4.0 & 155 &   73.3  \\
         ProxylessNAS\cite{cai2018proxylessnas} & 4.1 & 320 &  74.6  \\ 
         FBNet-B\cite{wu2019fbnet} & 4.5 & 295 &  74.1  \\ 
         IGCV3-D\cite{sun2018igcv3} & 7.2 & 610 &  74.6  \\ 
         GhostNet-1.3\cite{han2019ghostnet} & 7.3 & 226 &  75.7  \\
         EfficientNet-b0\cite{tan2019efficientnet} & 5.3 & 390 &  76.3   \\
         \midrule
         \nameofnet{}{}-1.0 & 3.4 & 300 &  74.02   \\ 
         \nameofnet{}{}-1.0$^\dagger$ & 3.94 & 330 &   76.05  \\
         \nameofnet{}{}-1.1$^\dagger$ & $\bold{4.28}$ & 420 &   $\bold{76.7}$  \\
        \bottomrule[1pt]
        \end{tabular}
       
        \label{tab:cross_comparisons}
\end{table}

\myPara{Effect of identity tensor multiplier}
Here, we investigate how the identity tensor multiplier (Sec.~\ref{sec:full_arch}) would
trades-off the model accuracy and latency. We use pytorch to generate the model and run it on Google Pixel 4XL. For each model, we measure the average inference time of 10 images as the final inference latency.
%
As shown in Table~\ref{tab:branch_multiplier}, the reduction
of the multiplier has subtle impacts on the classification
accuracy.
When half of the identity representations are removed,
the performance has no drop but the latency is improved.
When the multiplier is set to $1/6$, the performance 
decreases by 0.34\% from 74.02\% to 73.68\%, but with further improvement in terms of latency.
This indicates that introducing such a hyper-parameter
does matter for balancing the model performance and latency.


\begin{table}[t]
    \centering
    \small
    \setlength\tabcolsep{1.4mm}
    \renewcommand{\arraystretch}{1.1}
     \caption{Model performance and latency comparisons
        with different identity tensor multipliers.
        As can be seen, the latency can be improved by using
        lower identity tensor multipliers with only
        negligible sacrifice on the classification accuracy.}
    \begin{tabular}{cccccccc} \toprule[1pt] 
        No. & Models & Tensor multiplier & Param. (M) & Top-1 Acc. (\%) & Latency (ms) \\ \midrule[1pt]
        1 & \nameofnet{}  & 1.0 & 3.4 &  74.02 & 211  \\
        2 & \nameofnet{}  & $1/2$ & 3.4 &  74.09 & 196 \\
        3 & \nameofnet{}  & $1/3$ & 3.4 &  73.91 & 195 \\ 
        4 & \nameofnet{}  & $1/6$ & 3.4 &  73.68 & 188 \\
        \bottomrule[1pt]
        \end{tabular}
        \label{tab:branch_multiplier}
\end{table}

\subsection{Application for Object Detection}

To explore the transferable capability of the proposed approach against MobileNetV2, in this subsection, we apply our classification model to the object detection task as pretrained models.
We use both the proposed network and MobileNetv2 as feature extractors and report results on the Pascal VOC 2007 test set \cite{everingham2015pascal} following \cite{liu2016ssd} using
SSDLite \cite{sandler2018mobilenetv2,liu2016ssd}.
Similar to \cite{sandler2018mobilenetv2}, the first and second layers of SSDLite are connected to the last pointwise convolution layer with output stride of 16 and 32, respectively.
The rest of SSDLite layers are attached on top of the last convolutional layer with output stride of 32.
During training, we use a batch size of 24 and all the models
are trained for 240,000 iterations.
For more detailed settings, readers can refer to
\cite{sandler2018mobilenetv2,liu2016ssd}.

In Table~\ref{tab:detection}, we show the results when different backbone networks are used.
Obviously, with the nearly the same number of parameters and
computation, SSDLite with our backbone improves the one with
MobileNetV2 by nearly 1\%.
This demonstrates that the proposed network has better
transferable capability compared to MobileNetV2.

\begin{table}[t]
    \centering
    \small
    \setlength\tabcolsep{1.4mm}
    \renewcommand{\arraystretch}{1.1}
    \caption{Detection results on the Pascal VOC 2007 test set. As can be seen, using the same SSDLite320 detector,
    replacing the MobileNetV2 backbone with our network achieves better results in terms of mAP. Note that the multipliers of both MobileNetV2 and our network are set to 1.0.}
    \begin{tabular}{cccccccc} \toprule[1pt] 
        No. & Method & Backbone & Param. (M) & M-Adds (B) & mAP (\%) \\ \midrule[1pt]
        1 & SSD300  & VGG \cite{simonyan2014very}  & 36.1 & 35.2 & 77.2  \\
        2 & SSDLite320  & MobileNetV2 \cite{sandler2018mobilenetv2} & 4.3 & 0.8  & 71.7 \\
        3 & SSDLite320  & \nameofnet{} & 4.3 & 0.8  & 72.6 \\
        \bottomrule[1.0pt]
        \end{tabular}
        \vspace{2pt}
        
        \label{tab:detection}
\end{table}

\subsection{Improving Architecture Search as Super-operators}
It has been verified in previous subsections that our proposed \nameofblock{}  is more effective than the inverted residual block in both the classification task and the object detection task.
From a holistic perspective, we can also regard a residual
block as a `super' operator with more powerful transformation
power than a regular convolutional operator.
To further investigate the superiority of the
proposed \nameofblock{} over the inverted residual block,
we separately add it into the search space
of the differentiable searching algorithm (DARTS) \cite{liu2018darts} 
to see the network performance after architecture search
and report the corresponding results on CIFAR-10 dataset.
As shown in Table~\ref{tab:darts_i2rnet}, by adding our
\nameofblock{} as an new operator into the DARTS search space
without changing the cell structure, the resulting model
achieves higher accuracy than the model with the original
DARTS search space with about 25\% parameter reduction.
However, the searched model with the inverted residual block
added in the search space decreases the original performance.
This demonstrates that our proposed \nameofblock{} can generate more expressive representations than the inverted residual block
and can also be used in architecture search algorithms as a 
kind of `super' operator.
For more details on the searched cell structure, please
 refer to our supplementary materials.


%
\begin{table}[t]
    \centering
    \small
    \setlength\tabcolsep{1.5mm}
    \renewcommand{\arraystretch}{1.1}
    \caption{Results produced by different network architectures searched by DARTS \cite{liu2018darts}. For Lines 2 and 3, we separately add the inverted residual (IR) block and our \nameofblock{} into the original search space of DARTS.
    We report results on CIFAR-10 dataset as in \cite{liu2018darts}.}
    \begin{tabular}{cccccccc} \toprule[1pt] 
        No. & Search Space & Test Error (\%) & Param. (M) & Search Method &   \#Operators \\ \midrule[1pt]
        1 & DARTS original  & 3.11 & 3.25 &  gradient based & 7  \\
        2 & DARTS + IR Block  & 3.26 & 3.29 & gradient based  & 8 \\
        3 & DARTS + \nameofblock{}  & 2.98 & 2.45 & gradient based  & 8 \\
        \bottomrule[1.0pt]
        \end{tabular}
        \vspace{2pt}
       
        \label{tab:darts_i2rnet}
\end{table}


\section{Conclusions}
In this paper, we deeply analyze the design rules and
shortcomings of the previous inverted residual block.
Based on the analysis, we propose to reverse the thought 
of adding shortcut connections between low-dimensional
representations and present a novel building block, called
the \nameofblock{}, that connects high-dimensional representations instead.
We furthermore break through the tradition of previous
residual blocks using one spatial convolution in each
and emphasize the importance of using one more such convolution.
Experiments in both classification, object detection, and
neural architecture search demonstrate the effectiveness 
of the proposed \nameofblock{} and its potential to be used in more contexts.

\paragraph{Acknowledgement}
Jiashi Feng was partially supported by MOE Tier 2 MOE2017-T2-2-151, NUS\_ECRA\_FY17\_P08, AISG-100E-2019-035.

%
%
\bibliographystyle{splncs04}
\bibliography{egbib}
\clearpage

\appendix

\begin{figure}[h]
    \centering
    \includegraphics[width=\linewidth]{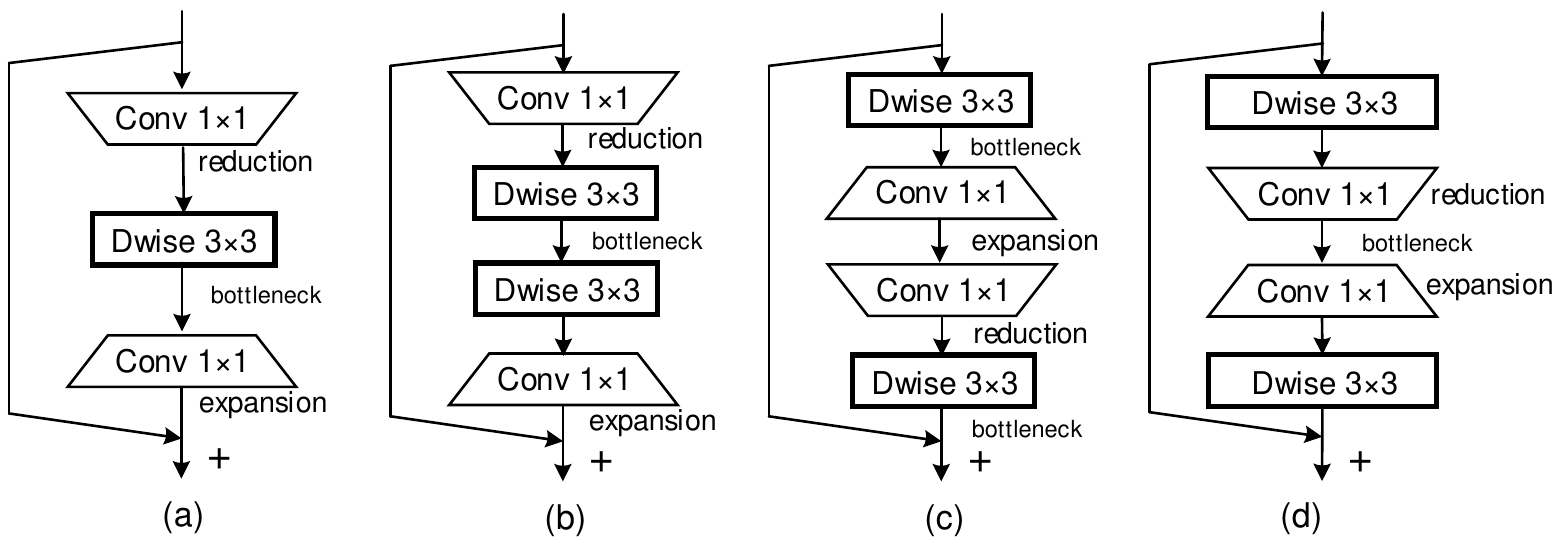}
    \caption{
    Conceptual diagram for variants of our proposed \nameofblock{}.
    (a) Bottleneck structure with standard convolutions replaced by depthwise convolutions. 
    (b) Structure in (a) with one more depthwise convolution in the bottleneck.
    (c) Switching the positions of the two pointwise convolutions in our \nameofblock{}.
    (d) Our proposed \nameofblock{}.
    }
    \label{fig:variants}
\end{figure}

\section{ Variants of the Proposed Sandglass Block}

In this section, we introduce and compare the variants of our proposed \nameofblock{}, which
are shown in Figures~\ref{fig:variants} (a-c).
The corresponding results are listed in
Table~\ref{tab:variants}.

\begin{enumerate}
    \item The first variant (Figure~\ref{fig:variants}(a)) is built from direct modification of the classic bottleneck structure \cite{he2016deep} by replacing the standard $3\times3$ convolution with a $3\times3$ depthwise convolution. 
    From the result, we can observe performance drop of about 5\% compared to our \nameofblock. 
    We argue this is mostly because the depthwise convolution is conducted in the bottleneck with a low-dimensional feature space
    and hence cannot capture enough spatial information,
    leading to much worse performance compared to 
    our proposed \nameofblock{} (Figure~\ref{fig:variants}(d)).

\item The second variant (Figure~\ref{fig:variants}(b)) is derived from the first variant, 
but differently we add another $3\times 3$ depthwise convolution in the bottleneck. 
As can be seen, the top-1 accuracy improves by more than 1\% compared to the structure shown in Figure~\ref{fig:variants}(a). 
This indicates encoding more spatial information indeed helps.
This phenomenon can also be observed by comparing Figure~\ref{fig:variants}(b) and Figure~\ref{fig:variants}(d) (70.11 \emph{v.s.} 74.02).

\item The third variant (Figure~\ref{fig:variants}(c)) is based on the original inverted residual block \cite{sandler2018mobilenetv2}.
We move the depthwise convolution from the high-dimensional feature space to the bottleneck positions with less feature channels. 
Compared with Figure~\ref{fig:variants}(b), this variant in Figure~\ref{fig:variants}(c) has a comparable number of learnable parameters and more computational cost but worse performance (69.26 \emph{v.s.} 70.11). 
This also means building shortcuts between high-dimensional representations is more beneficial to the network performance.
\end{enumerate}

As shown in Table~\ref{tab:variants}, our proposed \nameofblock{} achieves much better results than all the three variants.
The performance improvements can be explained by the two rules that we have presented in the main paper: 
(1) adding shortcut connections between high-dimensional representations,
and (2) performing the depthwise convolution in high-dimensional feature space.
%
%
Our experiments also indicate that bottleneck structure is suitable for mobile networks and it can work better than the inverted residual block.

\begin{table}[t]
    \centering
    \small
    \setlength\tabcolsep{1.4mm}
    \renewcommand{\arraystretch}{1.1}
    \caption{Performance of different variants of our proposed \nameofblock{}
    shown in Figure~\ref{fig:variants}.}
    \begin{tabular}{cccccccc} \toprule[1pt] 
        Block structure & \#Dwise convs & Param. (M) & M-Adds (M) & Top-1 Acc. (\%)  \\ \midrule[0.6pt]
        MobileNetV2 & 1  & 3.5 & 300 &  72.3   \\ \midrule[0.6pt]
        Figure~\ref{fig:variants}(a) & 1 & 3.4 & 240 &  68.90 \\ 
        Figure~\ref{fig:variants}(b) & 2 & 3.4 & 250 &  70.11  \\ 
        Figure~\ref{fig:variants}(c) & 2 & 3.5 & 300 &  69.26  \\ 
        Figure~\ref{fig:variants}(d) & 2 & 3.5 & 300 &  74.02  \\ 
        \bottomrule[1pt]
        \end{tabular}
       
        \label{tab:variants}
\end{table}

\begin{table}[h]
    \centering
    \small
    \setlength\tabcolsep{1.5mm}
    \renewcommand{\arraystretch}{1.1}
    \caption{Results produced by different network architectures
    searched by DARTS \cite{liu2018darts}. For Lines 2 and 3, we
    separately add the inverted residual (IR) block and our
    \nameofblock{} into the original search space of DARTS.
    We report results on CIFAR-10 dataset as in \cite{liu2018darts}.}
    \begin{tabular}{cccccccc} \toprule[1pt] 
        No. & Search Space & Test Error (\%) & Param. (M) & Search Method &   \#Operators \\ \midrule[1pt]
        1 & DARTS original  & 3.11 & 3.25 &  gradient based & 7  \\
        2 & DARTS + IR Block  & 3.26 & 3.29 & gradient based  & 7 \\
        3 & DARTS + \nameofblock{}  & 2.98 & 2.45 & gradient based  & 7 \\
        \bottomrule[1.0pt]
        \end{tabular}
        \vspace{2pt}
       
        \label{tab:darts_results}
\end{table}

\section{Searched Architectures}
Following \cite{liu2018darts,zoph2018learning,cai2018proxylessnas}, we also search for a computation cell and use it as the basic building block for the final architecture. 
The searching space and algorithm are described in details as below.

\myPara{Search space} In our experiments, we use the   search space  from \cite{liu2018darts} as our baseline (denoted as \emph{original}), which includes the following operators:
\begin{itemize}
    \item Convolutional operations (ConvOp): regular convolution, dilated convolution, depthwise convolution;
    \item Convolution kernel size\footnote{For the inverted residual block and our \nameofblock, we only use a kernel size of $3\times3$ for depthwise convolutions.}: $3\times3$, $5\times5$, $7\times1$ followed by $1\times7$;
    \item Non-parametric operations: average pooling, max pooling, skip connection, None.
\end{itemize}
The results are reported in Line 1 of Table~\ref{tab:darts_results}.
To compare with the inverted residual block, we conduct architecture search within the following  two new search spaces.
\begin{itemize}
    \item \textbf{Original + IR block}: the original search space plus the inverted residual block as the depthwise separable convolution operation candidate.
    
    \item \textbf{Original + \nameofblock{}}: the original search space plus the \nameofblock{}. 
\end{itemize}
%
The corresponding results are reported in Lines 2-3 of Table~\ref{tab:darts_results}, respectively.
The zero (None) operation is also included to indicate the miss of the connections as used in \cite{liu2018darts}.

\myPara{Searching algorithm} As mentioned in the main
paper, we adopt the DARTS searching algorithm  \cite{liu2018darts} to search for the cell structure and set the number of nodes to 7 in each directed acyclic graph (DAG) of the cell.
During the searching process, we strictly follow the training policy and use the same hyper-parameters  as in \cite{liu2018darts} for a fair comparison.
%

%
%

%

\begin{figure}[t]
    \centering
    \small
    \includegraphics[width=\linewidth]{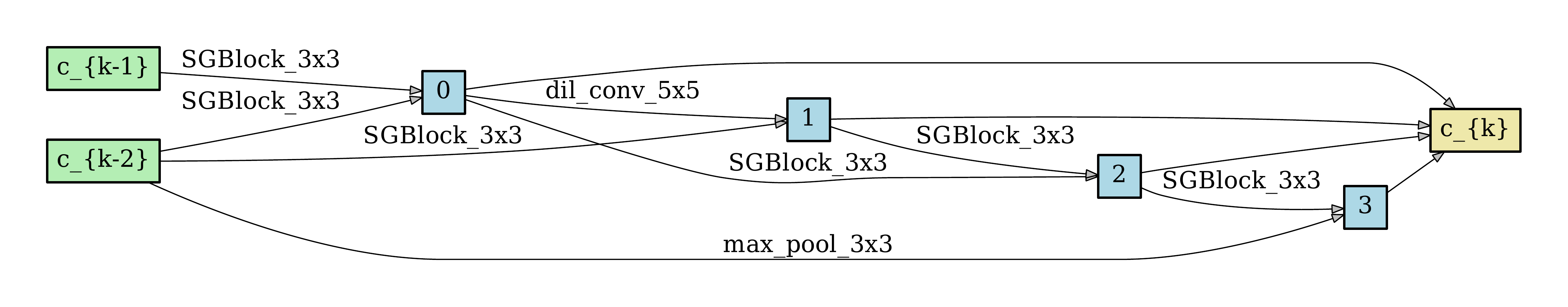}
    (a)
    \includegraphics[width=\linewidth]{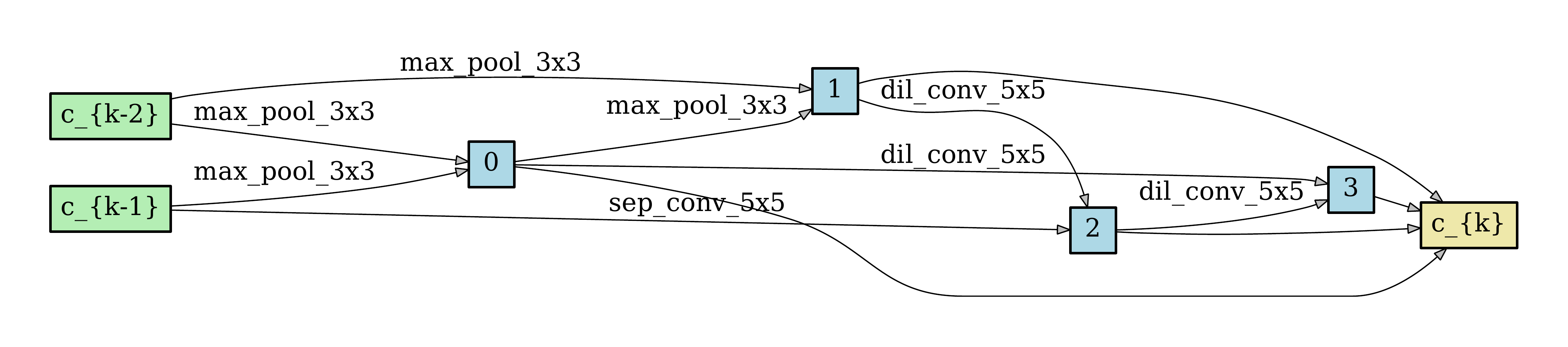}
    (b)
    \caption{
    Cell structures searched on CIFAR-10 with DARTS \cite{liu2018darts}. (a) Searched normal cell structure. (b) Searched reduction cell structure. `SGBlock' denotes our proposed \nameofblock{}. We use the same search space as used in \cite{liu2018darts} with only one more operator included, i.e. our proposed \nameofblock{}.
    }
    \label{fig:darts}
\end{figure}

\myPara{Architecture and results} The searched cell structures (including both the normal cell and the reduction cell) can be found in Figure~\ref{fig:darts}. 
As can be seen in Table~\ref{tab:darts_results},
adding the proposed \nameofblock{} into the original
search space can largely reduce the number of learnable parameters in the searched architecture
with improved classification performance on the CIFAR-10 dataset.
This again shows when combined with the searching algorithm, our proposed \nameofblock{} can be used to replace the original block to improve the performance.

\myPara{Conclusion and discussion} 
From the above results, we can observe that introducing appropriate super operators (\eg our \nameofblock{}) into the search space can bring better performance compared to using the original basic operators. 
We hope this experiment could benefit the development of architecture searching algorithms in the future.
\end{document}